\documentclass[sigconf, nonacm]{acmart}

\usepackage{graphicx}
\usepackage{booktabs} 

\usepackage{amsmath,amsfonts,bm}

\def\eqref#1{equation~\ref{#1}}

\def\1{\bm{1}}

\DeclareMathAlphabet{\mathsfit}{\encodingdefault}{\sfdefault}{m}{sl}
\SetMathAlphabet{\mathsfit}{bold}{\encodingdefault}{\sfdefault}{bx}{n}

\newcommand{\E}{\mathbb{E}}

\DeclareMathOperator*{\argmax}{arg\,max}
\DeclareMathOperator*{\argmin}{arg\,min}

\DeclareMathOperator{\sign}{sign}

\newcommand{\norm}[1]{\left\lVert#1\right\rVert}
\newcommand\inner[2]{\langle #1, #2 \rangle}
\newcommand{\logloss}[1]{\log(1+e^{#1})}
\newcommand{\linloss}{-yx^T\theta}
\newcommand{\advlinloss}{-yx^T\theta + c\norm{\theta}}
\newcommand{\advlinlossexample}{-yx^T\theta + c\norm{\theta}_{*}}
\newcommand{\ilinloss}{-y_ix_i^T\theta}
\newcommand{\rderiv}{-yx + c\frac{\theta}{\norm{\theta}}}
\newcommand{\rderivopt}{-yx + c\frac{\theta^*}{\norm{\theta^*}}}

\newcommand{\avgsum}{\frac{1}{n}\sum_{i=1}^n}
\newcommand{\gdavgsum}[2]{\frac{#1}{#2 n}\sum_{i=1}^n}
\newcommand{\hth}{h(\theta)}
\newcommand{\hthopt}{h(\theta^*)}
\newcommand{\rth}{r(\theta)}

\newcommand{\dhth}{h'(\theta)}
\newcommand{\drth}{r'(\theta)}
\newcommand{\drthopt}{r'(\theta^*)}

\newcommand{\clip}[2]{\min(1, \frac{#2}{\norm{#1}})#1}
\usepackage{multirow}

\usepackage{nicefrac}
\usepackage{amsmath}
\usepackage{amsfonts}
\usepackage{pifont}
\newcommand{\cmark}{\ding{51}}%
\newcommand{\xmark}{\ding{55}}%

\usepackage{caption}
\usepackage{subcaption}

\usepackage{hyperref}
\usepackage{cleveref}

\begin{document}

\pagestyle{fancy}
\fancyhf{}
\fancyhead[RE,RO]{Ongoing work. Appeared in PPML, 2021}
\fancyfoot[C]{\thepage}

\title{Learning to be adversarially robust and differentially private}

\author{Jamie Hayes, Borja Balle, M. Pawan Kumar}
\affiliation{DeepMind}

\begin{abstract}

We study the difficulties in learning that arise from robust and differentially private optimization.
We first study convergence of gradient descent based adversarial training with differential privacy, taking a simple binary classification task on linearly separable data as an illustrative example.
We compare the gap between adversarial and nominal risk in both private and non-private settings, showing that the data dimensionality dependent term introduced by private optimization compounds the difficulties of learning a robust model.
After this, we discuss what parts of adversarial training and differential privacy hurt optimization, identifying that the size of adversarial perturbation and clipping norm in differential privacy both increase the curvature of the loss landscape, implying poorer generalization performance.
\end{abstract}

\maketitle

\section{Convergence rates of the logistic loss on a linearly separable problem with gradient descent}

\begin{table*}
\small
\centering
\caption{Convergence rates of empirical nominal and adversarial risk of logistic loss in non-private and private settings.}
\label{tab:logloss_cr}
\begin{tabular}{cccl}
\toprule
    \multicolumn{2}{c}{\textbf{Optimizer}}              &
    \multirow{ 2}{*}{\textbf{Loss}}         &
    \multirow{ 2}{*}{\textbf{Bound}}  \\
\cmidrule{1-2}
\textbf{Robust} & \textbf{Private} & \\
\midrule
\xmark & \xmark                  &
$L(\theta^t)$ &
$\frac{8-\eta}{8 t\eta}\bigg((\frac{\log t}{\gamma})^2 + 1\bigg) + (\frac{8-\eta}{4})\log(\frac{t+1}{t})$           \\
\xmark    & \cmark       &
$\mathbb{E}[L(\theta^t)]$ &
$\frac{8-\eta}{8 t\eta}\bigg((\frac{\log t}{\gamma})^2 + 1 + d\sigma^2\bigg) + (\frac{8-\eta}{4})\log(\frac{t+1}{t}) +  \eta d\sigma^2$            \\
\cmark     & \xmark       &
$L_a(\theta^t)$ &
$\big(\frac{8-\eta(1+c)^2}{8 t\eta} - \frac{c}{t\gamma}\big) \bigg((\frac{\log t}{\gamma - c})^2 + (1+c)^2\bigg) + \big(\frac{8-\eta(1+c)^2}{4} -\frac{2c}{\gamma}\big)\log(\frac{t+1}{t})$          \\
 \cmark &  \cmark &
 $\mathbb{E}[L_a(\theta^t)]$ &
 $\big(\frac{8-\eta(1+c)^2}{8 t\eta} - \frac{c}{t\gamma}\big)\bigg(  (\frac{\log t}{\gamma - c})^2 + (1+c)^2 + d\sigma^2 \bigg) + \big(\frac{8-\eta(1+c)^2}{4} -\frac{2c}{\gamma}\big)\log(\frac{t+1}{t}) +  \eta d\sigma^2$ \\
\bottomrule
\end{tabular}
\end{table*}

We begin by analyzing logistic regression as a motivating example.

Let $\ell(\cdot, (\cdot, \cdot)):\mathcal{W}\times (\mathcal{X}, \mathcal{Y}) \rightarrow \mathbb{R}$ be the logistic loss $\ell(\theta; x,y)=\logloss{\linloss}$. We assume $\mathcal{X}\subseteq \mathbb{R}^d$ and $\forall (x,y)\in(\mathcal{X}, \mathcal{Y})$, $\norm{x}\leq 1$ and $y\in \{\pm1\}$.
We further assume the data is linearly separable with margin $\gamma$, and
set $u=\argmax_{\norm{\theta}=1}\min_{i\in[n]}y_ix_i^T\theta$, the optimal hyperplane that classifies all $(x,y)$ correctly with margin at least $\gamma$.
We let $t\in\mathbb{N}$ denote the training step, and $\eta\in\mathbb{R}$ denote the learning rate. 
We will see that the difference in rate of convergence between non-robust and robust optimization grows with the dimensionality $d$ of the input space $\mathcal{X}$.

Following a similar analysis in \cite{li2019inductive}, we compute convergence rates for the loss under gradient descent with and without adversarial training and differential privacy.
Without robustness we optimize the loss $L(\theta) = \frac{1}{n}\sum_{i=1}^n \ell(\theta; x_i, y_i)$, while
in adversarial training we minimize the adversarial loss $L_a(\theta) = \frac{1}{n}\sum_{i=1}^n \ell_a(\theta; x_i, y_i)$, where $\ell_a(\theta) = \max_{\norm{\delta}\leq c} \logloss{-y(x+\delta)^T\theta} = \logloss{\advlinlossexample}$, where $c$ is the size of adversarial perturbations. 
Throughout our analysis we use the $\ell_2$ norm on $L_a$, however we also conduct experiments with the $\ell_{\infty}$ norm.
Under differential privacy the gradient at each time step is perturbed by Gaussian noise with variance $\sigma$ --- the privacy guarantees can be computed based on the amount of noise, number of iterations and Lipschitz constant of the loss in a standard way \cite{bassily2014private}.

\Cref{tab:logloss_cr} gives the converge rates in each of our four possible optimization settings, with proofs given in \cref{sec: conv_rates_proofs}.
\footnote{We stress that we do not claim our bounds are tight. For example, we should be able to improve the bound in the non-private non-robust setting by a factor of $\log t$ by appealing to Theorem 3.3 in \cite{bubeck2014convex}.} 
We observe that optimizing the adversarial loss introduces a dependence on the adversarial budget $c$ in the convergence bound, while making the optimizer differentially private introduces terms $d \sigma^2$ depending on the dimension and amount of noise.
Combining adversarially robust and private optimization introduces the two modifications on the bound almost independently -- the only cross-contribution is through a \emph{negative} term.
In general, two factors dominate the rate of convergence in robust and private optimization: (i) the dimensionality of data $d$; and (ii) the size of the adversarial budget $c$ with respect to the size of the margin $\gamma$.

We discuss some interesting observations that arise from considering these different optimization settings.

\begin{figure}[h]
  \centering
  \includegraphics[width=0.95\columnwidth]{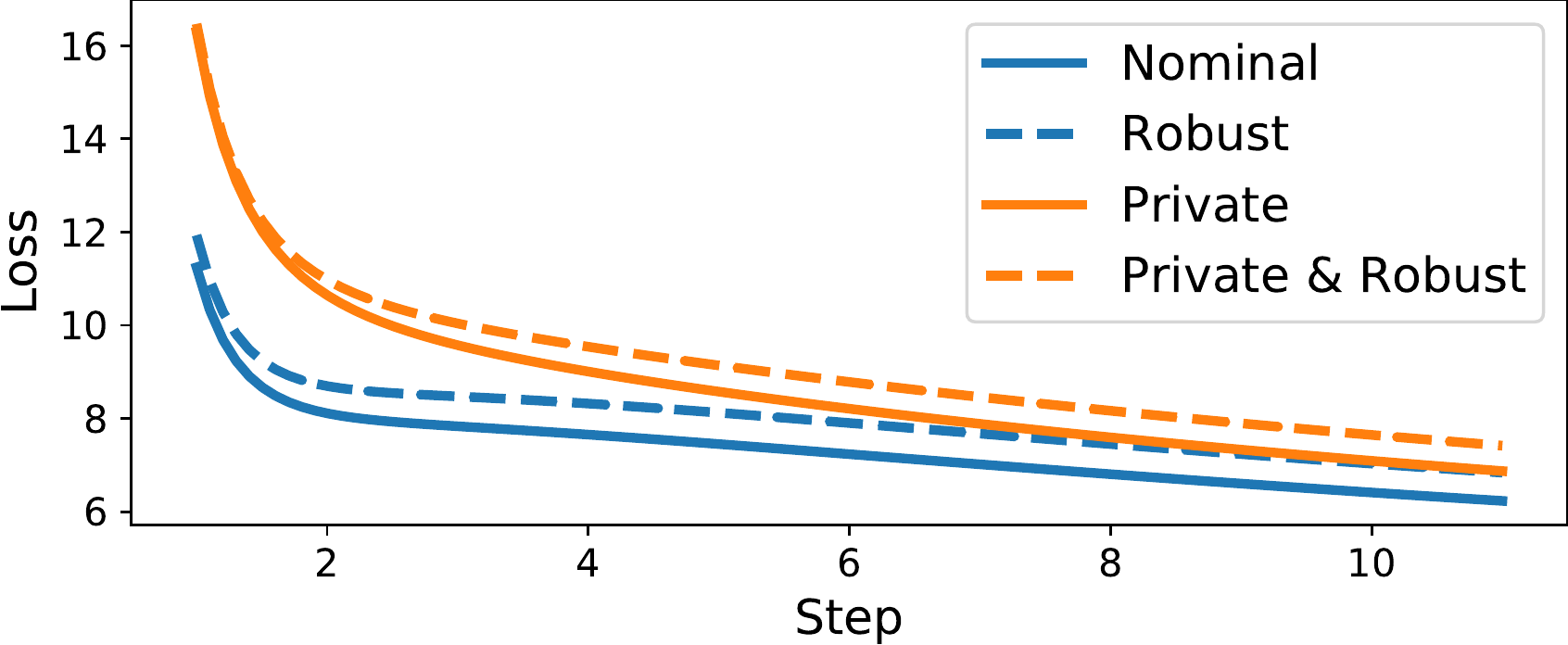}
  \captionof{figure}{Converge of logistic regression with $d=10$, $\sigma=0.25$, $\gamma=1.0$, $c=0.1$, $\eta=0.1$.}
  \label{fig:lr_example}
\end{figure}

\paragraph{Hierarchy of empirical risk bounds}
We inspect the hierarchy of empirical nominal and adversarial risk in non-private and private settings.
Firstly, empirical nominal risk is always smaller than empirical adversarial risk in either setting.
The non-private empirical adversarial risk can be smaller or larger than private empirical nominal risk. The position of overlap depends upon respective hyperparameters $c$ and $\sigma$, and data dependent parameters such as dimensionality. 
These effects are shown in \cref{fig:lr_example} for a simple binary classification task.

\begin{figure}[h]
  \centering
  \includegraphics[width=0.95\columnwidth]{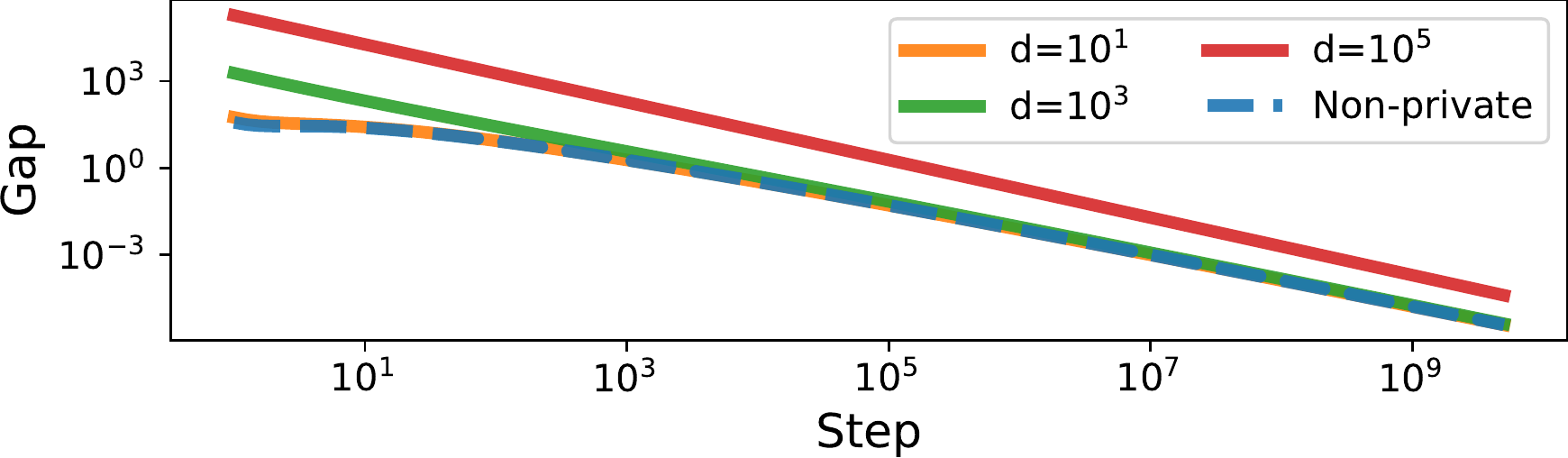}
  \captionof{figure}{Gap between bounds on empirical adversarial and nominal risk as a function as the number of optimization steps. We measure the gap in the non-private setting, and in the private setting for different values of $d$. We use the same parameters as described in \cref{fig:lr_example}.}
  \label{fig:lr_gap_example}
\end{figure}

\paragraph{The gap between empirical adversarial risk and nominal risk is approximately equal in non-private and private settings for small $d$}
Despite the aforementioned hierarchy, it is worth asking: does the gap between nominal and adversarial risk expand as we introduce privacy into the optimization process?
The answer is yes, however, this gap tends to zero over training given a sufficient number of steps. 
We measure the gap between adversarial and nominal risk throughout training in either the private or non-private setting.
\Cref{fig:lr_gap_example} shows that this gap tends to zero as $t\rightarrow\infty$, and that the gap increases in the private setting as $d$ increases. 

\begin{figure}[htp]
  \centering
  \includegraphics[width=0.95\columnwidth]{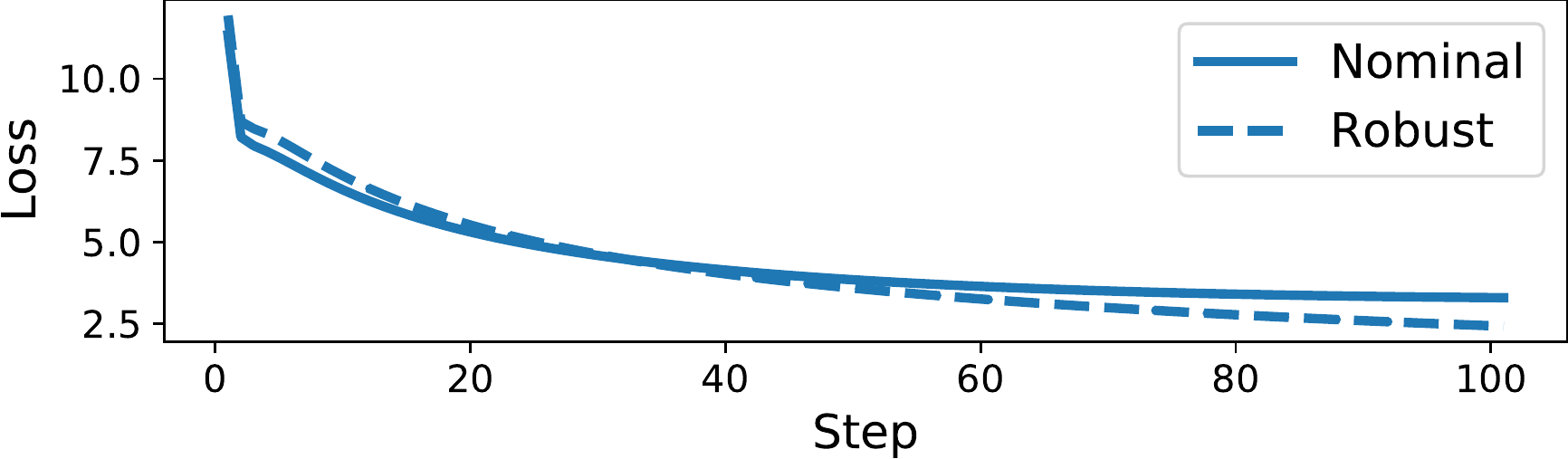}
  \captionof{figure}{Comparison of robust loss under gradient descent with and without adversarial training, using the same parameters as described in \cref{fig:lr_example}.}
  \label{fig:lr_row_compare_non_private}
\end{figure}

\paragraph{Empirical adversarial risk comparison}
It is worth noting that \cref{fig:lr_example} is not an apples-to-apples comparison because we are comparing standard loss ($L$) under standard training against robust loss ($L_a$) under adversarial training. 
We can compare with robust accuracy under standard training by noting that under gradient descent, we can upper bound on the robust loss by $L_a(\theta^t) \leq  \frac{8-\eta}{8 t\eta}\bigg(1 + (\frac{\log t}{\gamma})^2\bigg) + (\frac{8-\eta}{4})\log(\frac{t+1}{t}) + c\big(1 + \eta(t-1)\big)$ (a full derivation can be found in \cref{sec: conv_rates_proofs}).
We can then plot the robust loss under gradient descent and adversarial training + gradient descent in \cref{fig:lr_row_compare_non_private}, where the benefits of adversarial training are clear to see.

\paragraph{Hierarchy of convergence on non-linear models}

\begin{figure}[htp]
  \centering
  \includegraphics[width=0.95\columnwidth]{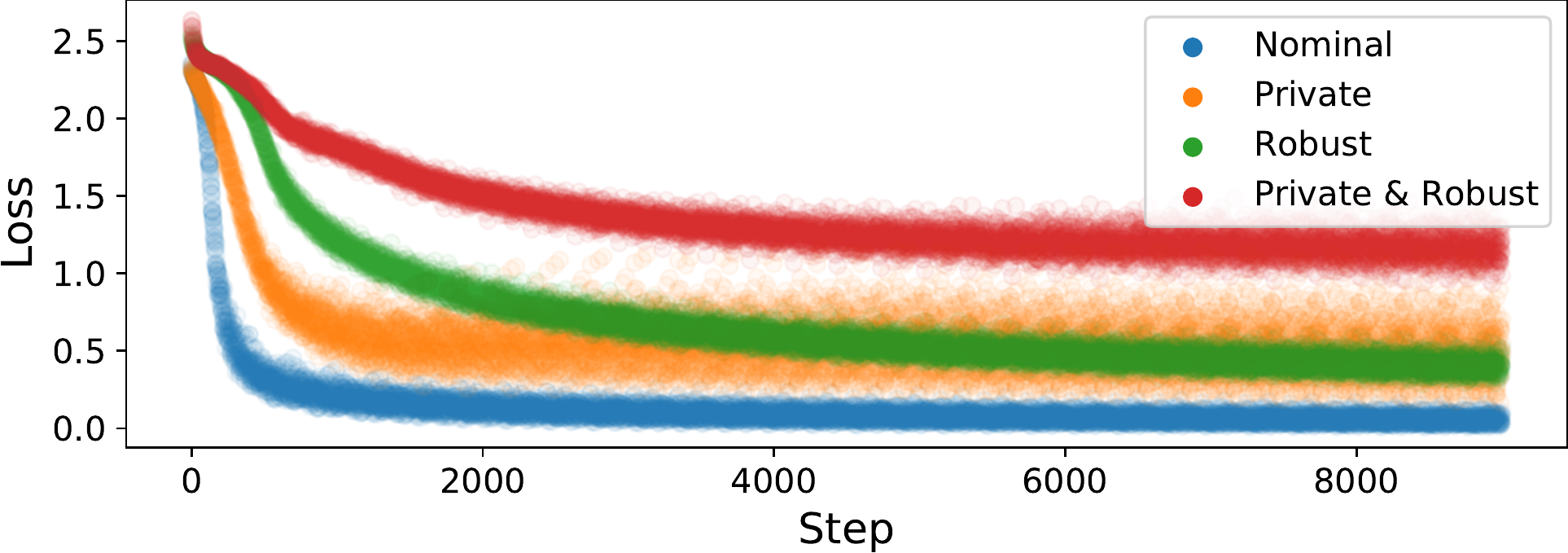}
  \captionof{figure}{Training loss of a three-layer CNN on MNIST.}
  \label{fig:mnist_loss}
\end{figure}

We now evaluate how well our bounds on logistic regression match the empirically observed convergence rates of non-linear models. 
We train a three layer convolutional neural network on the MNIST dataset, and evaluate canonical measures as introduced previously.
Firstly, \cref{fig:mnist_loss} shows the convergence hierarchy as previously introduced is preserved in practice; nominal training has the lowest loss while private and robust optimization incurs the largest loss over training. 
Nearly identical observations were made for the test loss.

The nominal model has a test set accuracy of 99.03\%, the robust model a test set accuracy of 99.20\%, the private model a test set accuracy of 96.10\%, and the robust and private model a test accuracy of 94.43\%. 
The private models are $(\epsilon, \delta)$-DP, with $\epsilon=3.8$ and $\delta=10^{-5}$, where individual gradients were clipped to norm one (cf.\ \citet{abadi2016deep}).
The robust models were trained with $\ell_{\infty}$ perturbations of size $0.3$ using 40 steps of PGD. All models are trained with standard SGD.
Additionally, we train analogous models on CIFAR-10 using six layers with max-pooling and tanh activations, and observe similar descent curves.
The private models are $(\epsilon, \delta)$-DP, with $\epsilon=8$ and $\delta=10^{-5}$, where individual gradients were clipped to norm 0.1. The robust models are trained to be robust to $\ell_{\infty}$ perturbations of size $\nicefrac{8}{255}$ using 20 steps of PGD.

\paragraph{Gap in practice}

\begin{figure}[htp]
  \centering
  \includegraphics[width=0.95\columnwidth]{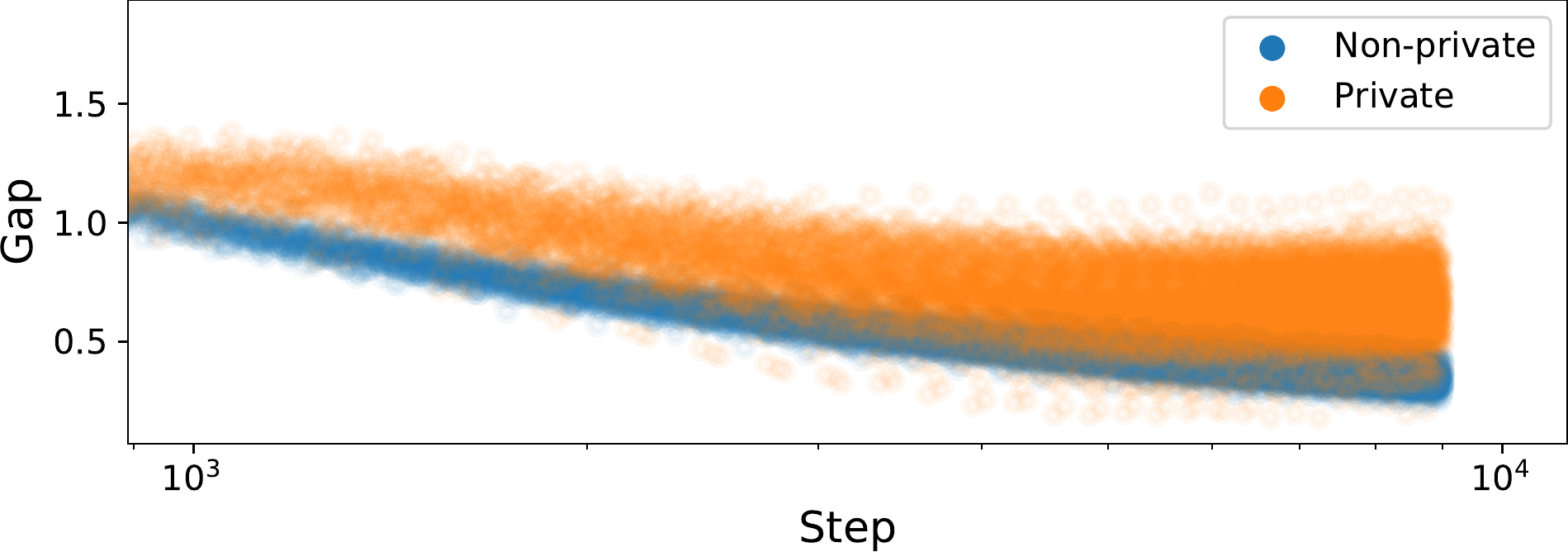}
  \captionof{figure}{Gap in loss between robust and non-robust optimization on MNIST.}
  \label{fig:mnist_loss_gaps}
\end{figure}

We now evaluate the gap between empirical adversarial and nominal risk on this non-linear model.
From \cref{fig:lr_gap_example}, we would expect that this gap shrinks as $t\rightarrow \infty$, with the gap being strictly larger in the private setting. This is precisely what we observe in \cref{fig:mnist_loss_gaps}.
Of course, we can no longer be confident the gap will tend to zero in this non-toy setting. 

\paragraph{Robustness gap}

\begin{figure*}[t]
\captionsetup[subfigure]{width=0.9\textwidth}
  \centering
\begin{subfigure}[t]{0.24\textwidth}
\centering
    \includegraphics[width=0.95\linewidth]{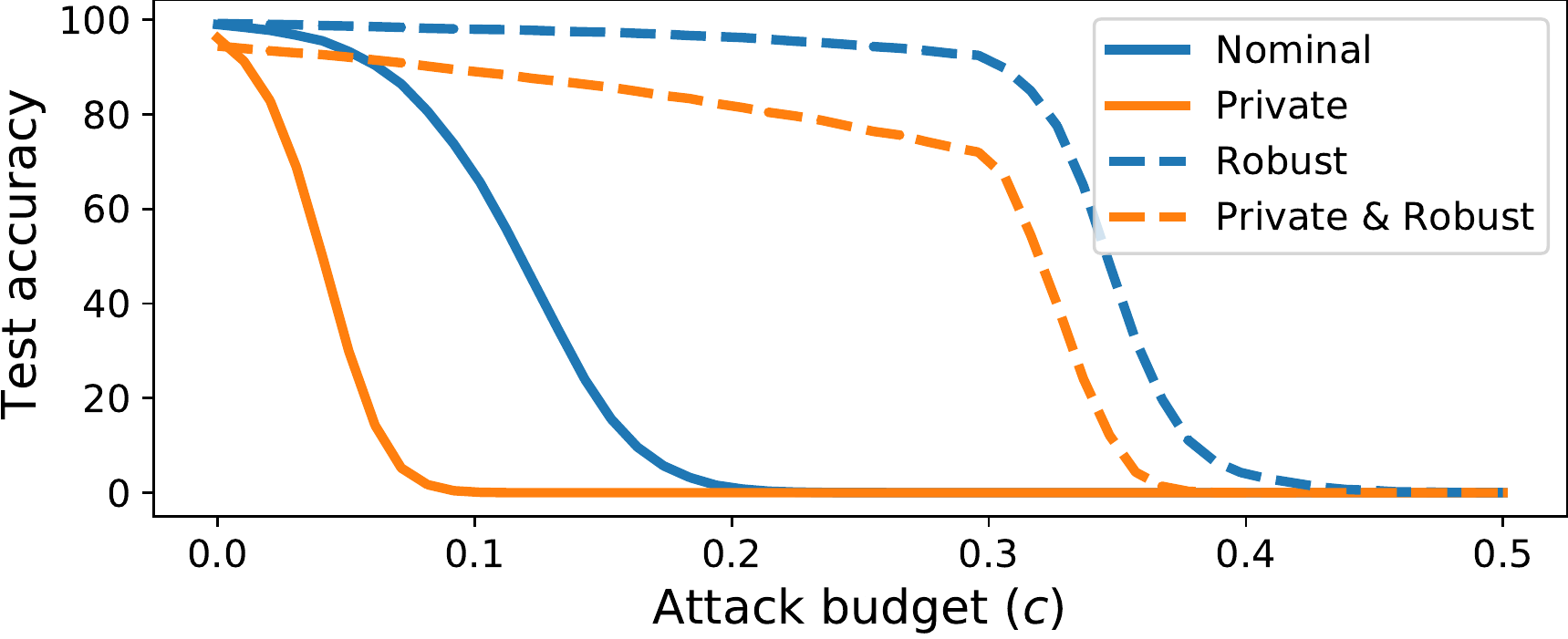}
  \caption{Robust accuracy on MNIST test set.}
  \label{fig:mnist_robustness}
\end{subfigure}%
\begin{subfigure}[t]{.24\textwidth}
\centering
    \includegraphics[width=0.95\linewidth]{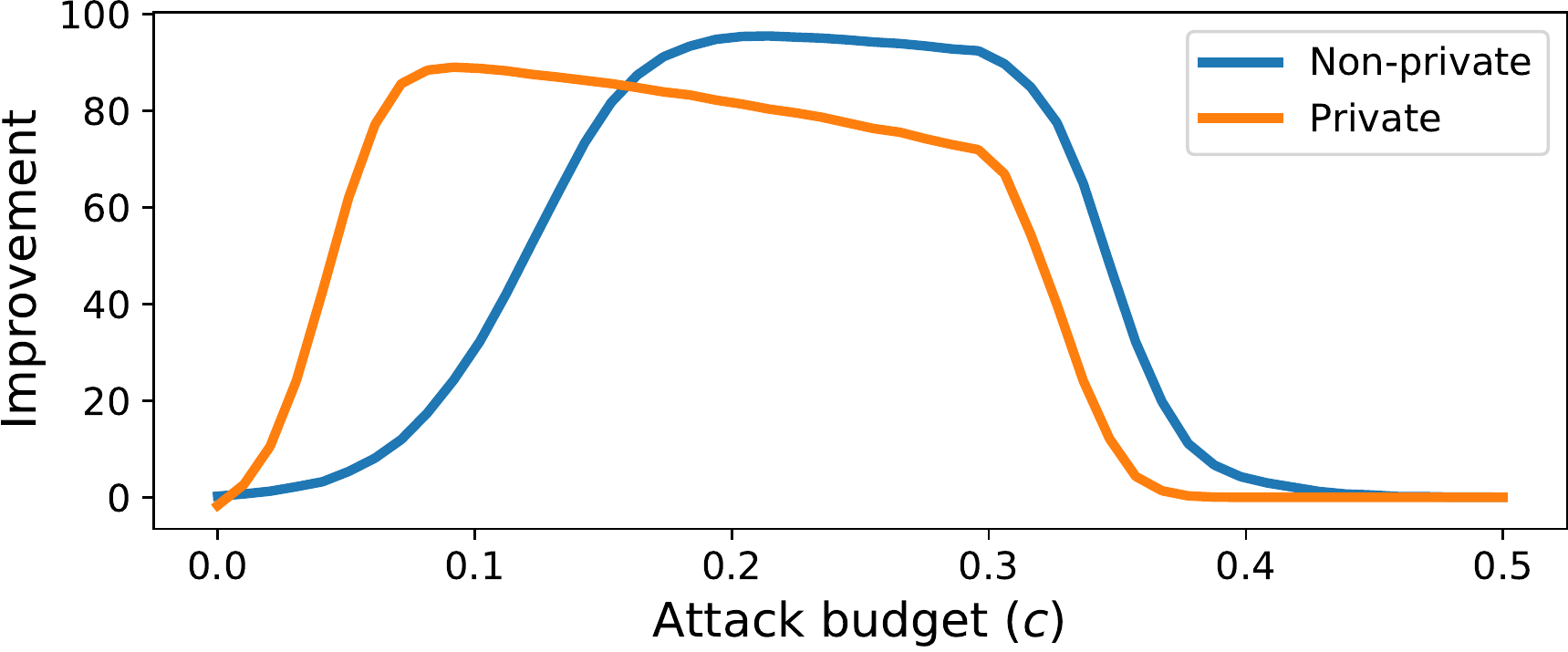}
  \caption{Improvement in robust accuracy with robust optimization on MNIST.}
  \label{fig:mnist_improvement}
\end{subfigure}%
\begin{subfigure}[t]{.24\textwidth}
\centering
    \includegraphics[width=0.95\linewidth]{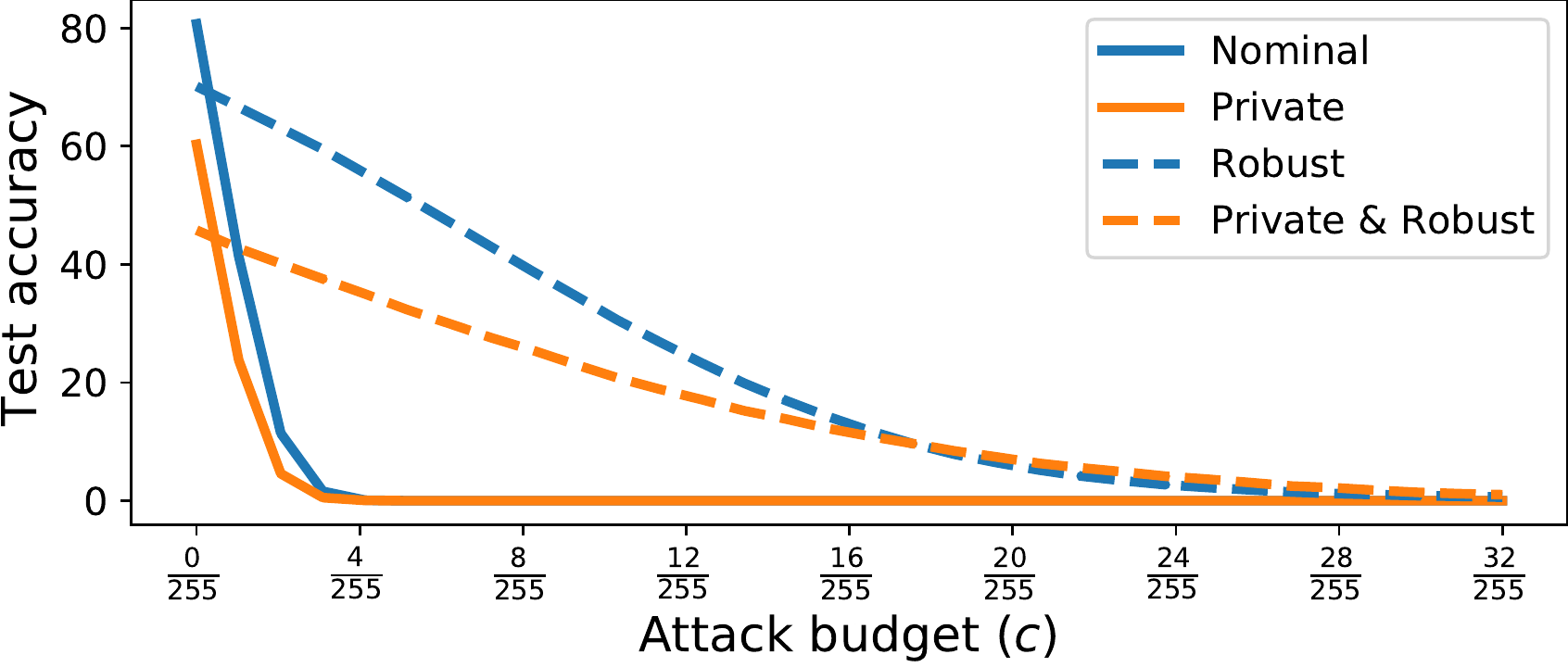}
  \caption{Robust accuracy on CIFAR-10 test set.}
  \label{fig:cifar10_robustness}
\end{subfigure}%
\begin{subfigure}[t]{.24\textwidth}
\centering
    \includegraphics[width=0.95\linewidth]{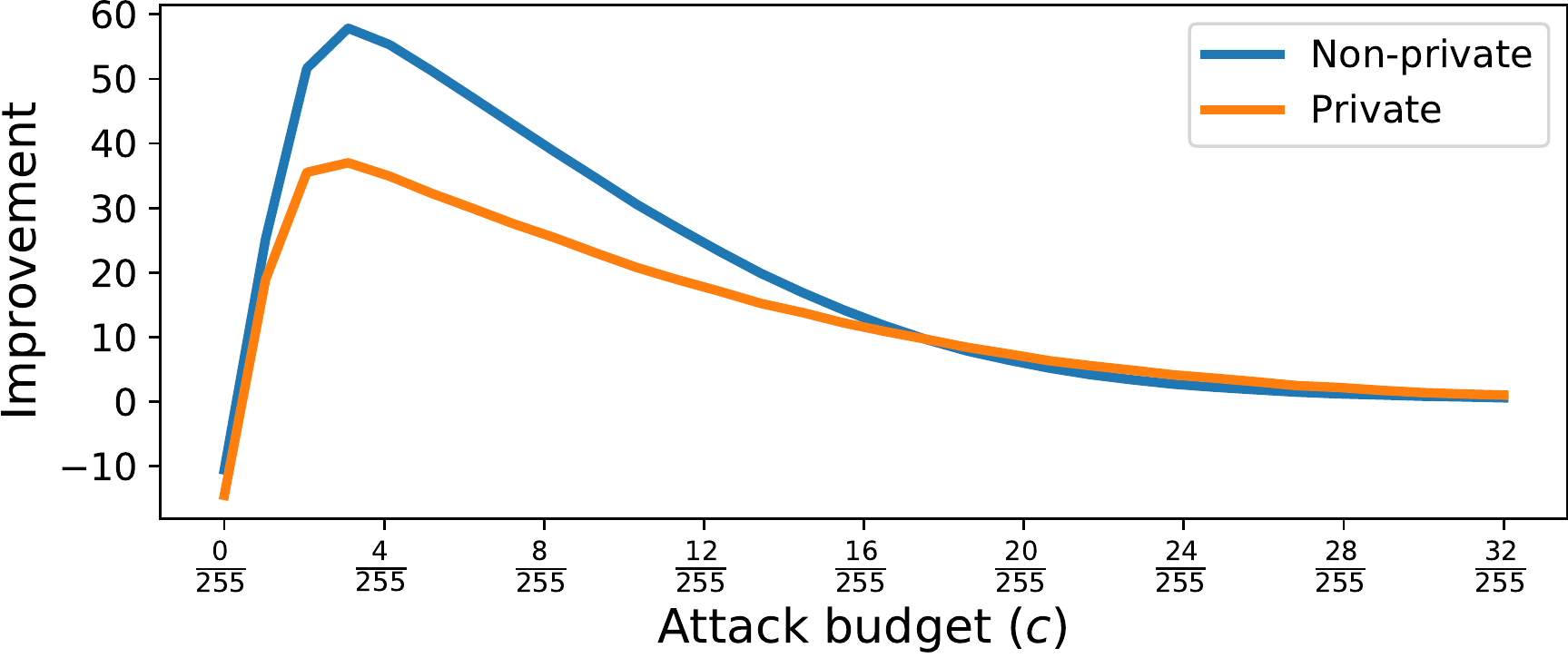}
  \caption{Improvement in robust accuracy with robust optimization on CIFAR-10.}
  \label{fig:cifar10_improvement}
\end{subfigure}%
\caption{Robustness comparison of MNIST and CIFAR-10. For \cref{fig:mnist_improvement} and \cref{fig:cifar10_improvement}, given an attack budget, $c$, we plot the difference in accuracy between a robust model and a nominal model, in either a non-private or private setting. We refer to this as the \emph{improvement} of the robust model over the nominal model.}
\label{fig:mnist_cifar10_compare}
\end{figure*}

In \cref{fig:mnist_robustness} and \cref{fig:cifar10_robustness}, we empirically evaluate the robustness of each model trained on MNIST and CIFAR-10, using 100 PGD steps. 
For both robust and non-robust optimized models, the private robustness curve resembles a shifted version of the non-private robustness curve. 

From \cref{fig:mnist_improvement}, we note the relative \emph{improvement} in robust accuracy when going from a non-robust model to a robust model is slightly higher in the non-private setting in comparison to the private setting. This implies that learning to be both robust and private is a more difficult learning task, and as a result robust accuracy suffers.
We also note, from \cref{fig:cifar10_robustness} and \cref{fig:cifar10_improvement}, the relative gap in improvement in robust accuracy between private and non-private optimization on CIFAR-10 models is larger than on MNIST. 
This is to be expected as our previous analysis suggested learning to be robust and private is more difficult for datasets of larger dimensions.

\section{The role of clipping, noise and adversarial budget}

\begin{figure*}[t]
\captionsetup[subfigure]{width=0.9\textwidth}
  \centering
\begin{subfigure}[t]{0.33\textwidth}
\centering
    \includegraphics[width=0.95\linewidth]{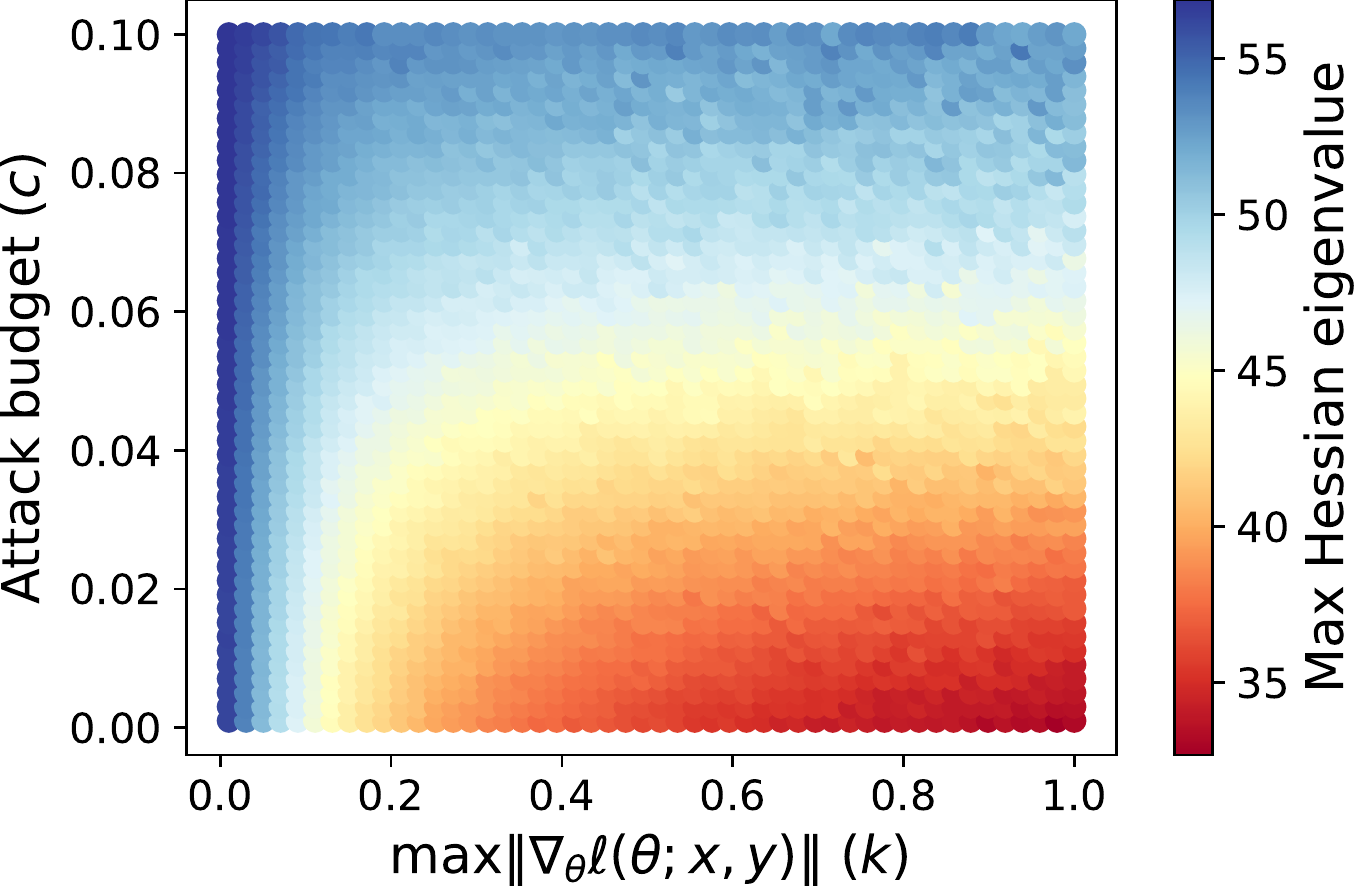}
  \caption{Maximum eigenvalue for different training attack budgets ($c$) against clipping thresholds ($k$).}
  \label{fig:log_reg_x_grad_max_norm_y_attack_epsilon_z_eigvals_25011747}
\end{subfigure}%
\begin{subfigure}[t]{.33\textwidth}
\centering
    \includegraphics[width=0.95\linewidth]{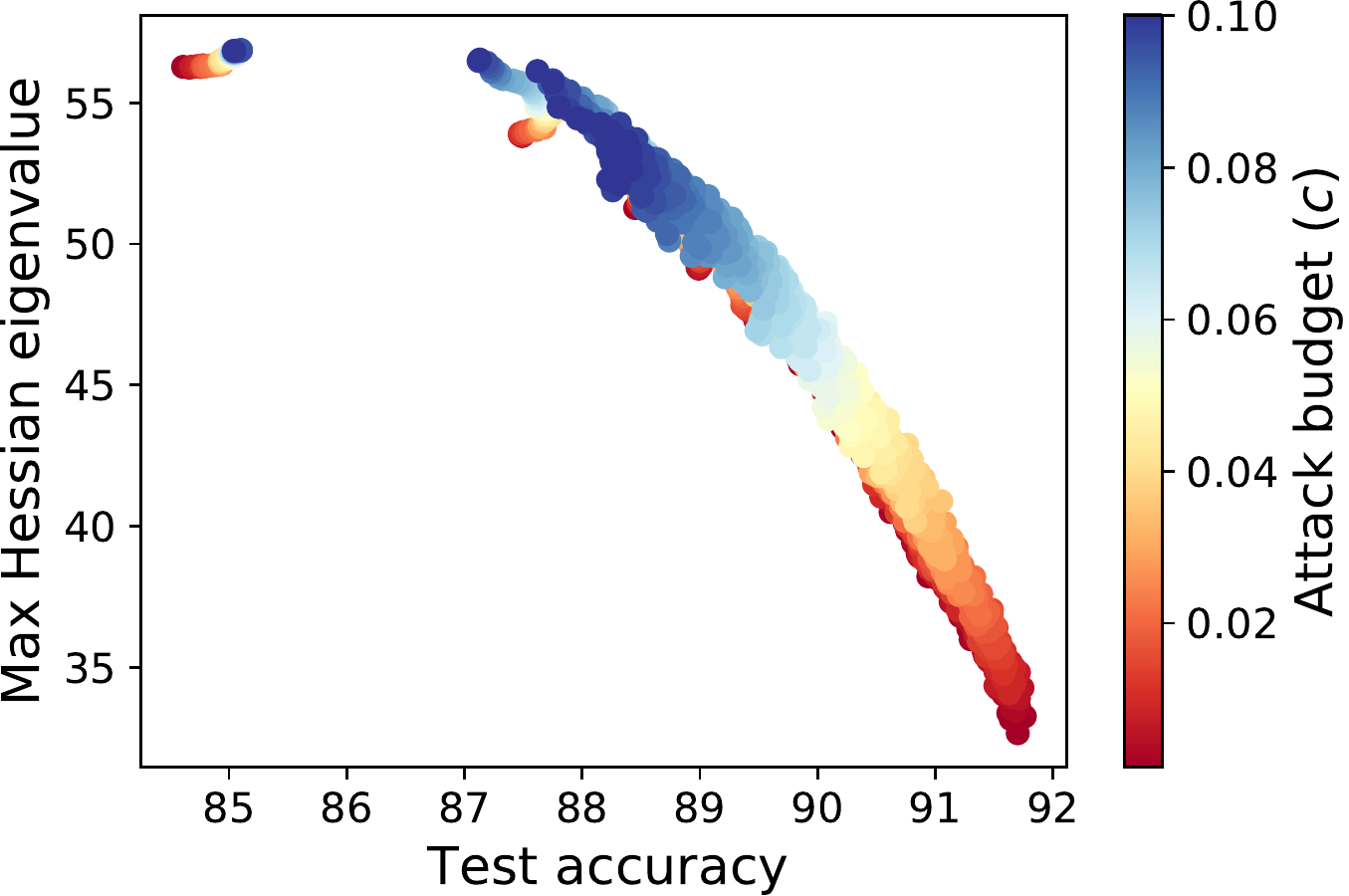}
  \caption{The correspondence between maximum eigenvalue and test accuracy for different training attack budgets ($c$).}
  \label{fig:log_reg_x_acc_eval_y_eigvals_z_attack_epsilon_25011747}
\end{subfigure}%
\begin{subfigure}[t]{.33\textwidth}
\centering
    \includegraphics[width=0.95\linewidth]{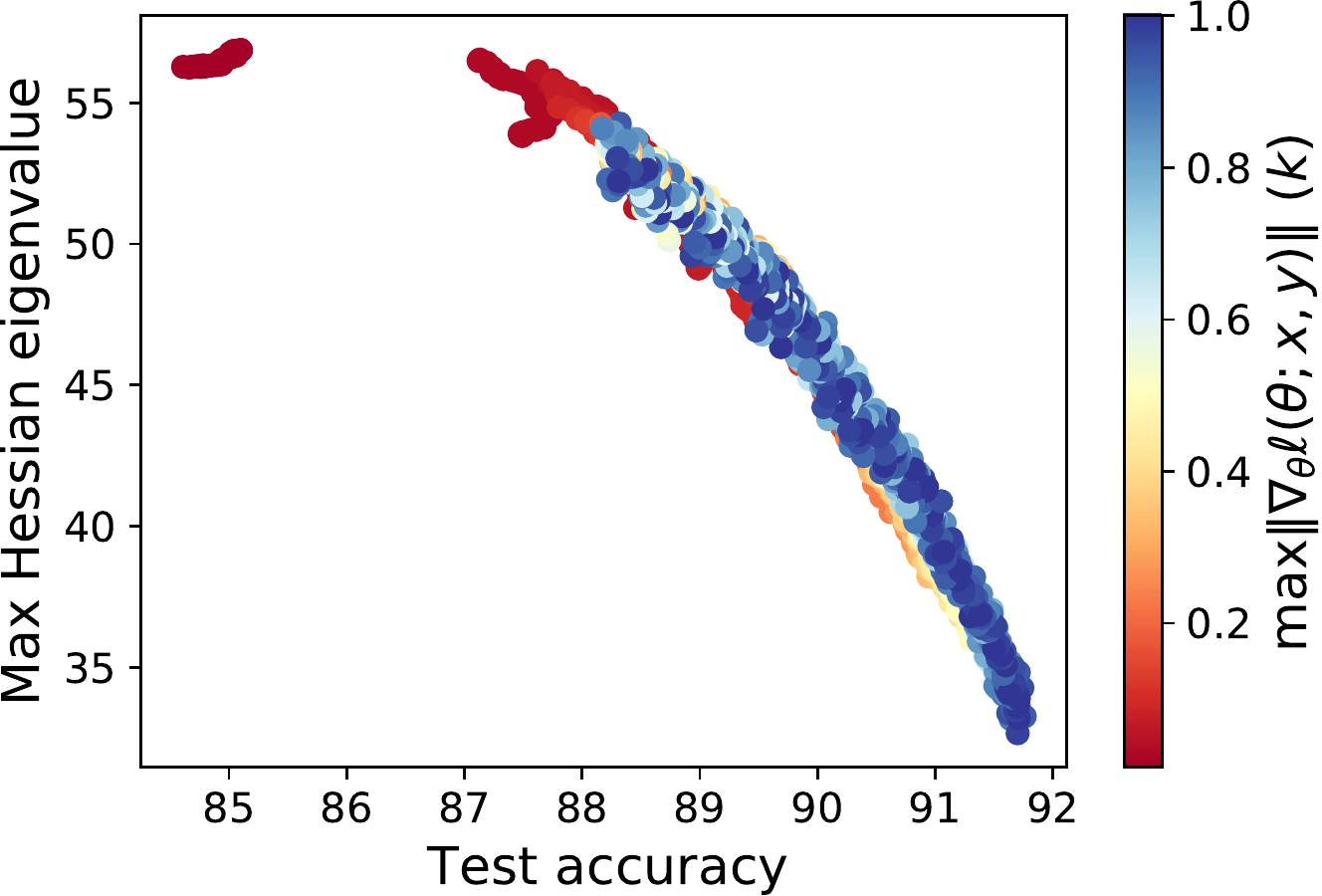}
  \caption{The correspondence between maximum eigenvalue and test accuracy for different clipping thresholds ($k$).}
  \label{fig:log_reg_x_acc_eval_y_eigvals_z_grad_max_norm_25011747}
\end{subfigure}
\caption{Robust and clipped logistic regression. Each point in the figures represents a different model trained to be robust to perturbations of size $c$ with gradients clipped to be smaller than $k$.}
\label{fig:log_reg_25011747}
\end{figure*}

\begin{figure*}[t]
\captionsetup[subfigure]{width=0.9\textwidth}
  \centering
\begin{subfigure}[t]{0.33\textwidth}
\centering
    \includegraphics[width=0.95\linewidth]{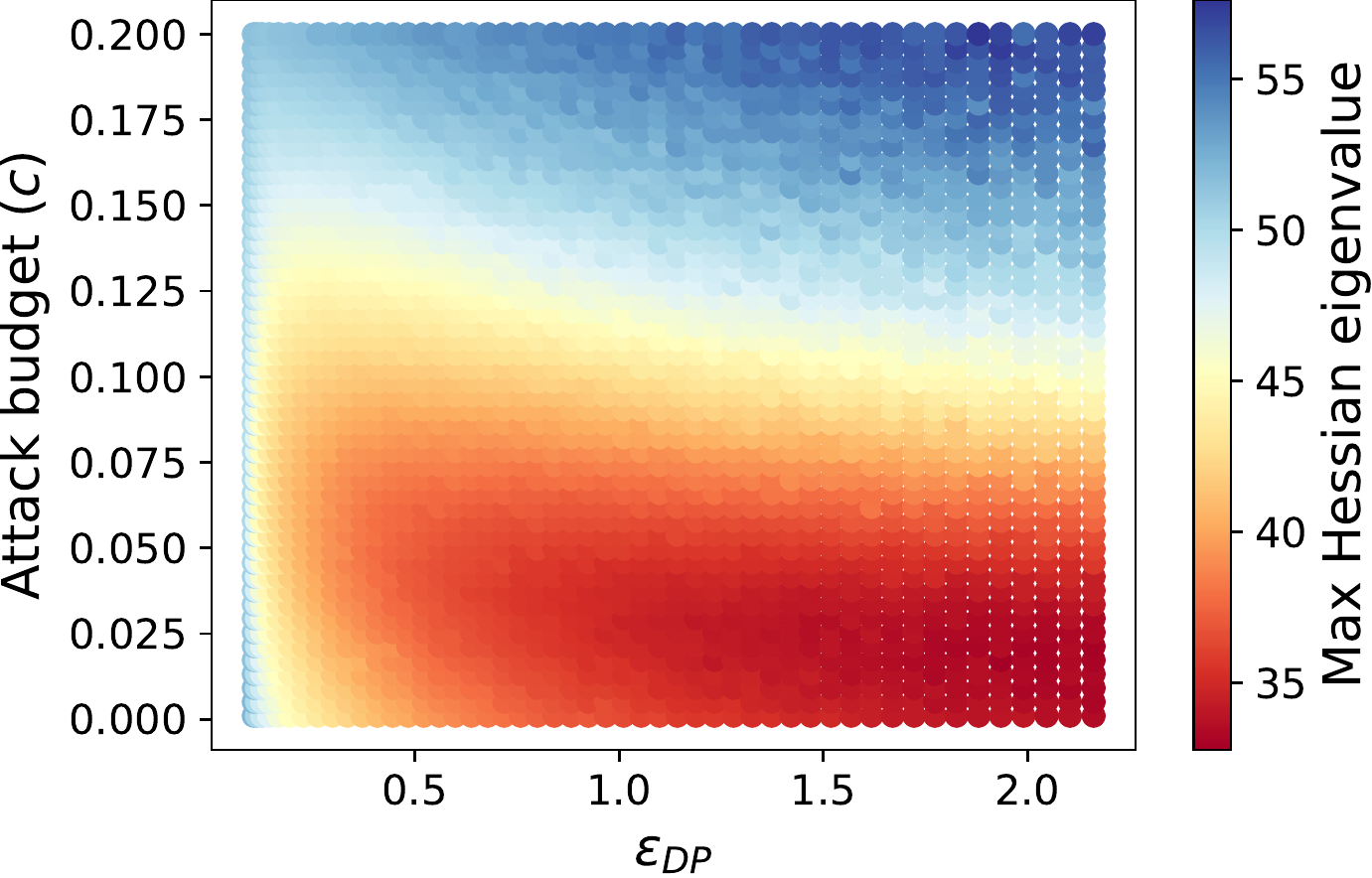}
  \caption{Maximum eigenvalue for different training attack budgets ($c$) against privacy guarantees ($\epsilon_{DP}$).}
  \label{fig:log_reg_x_dp_epsilon_y_attack_epsilon_z_eigvals_25056077}
\end{subfigure}%
\begin{subfigure}[t]{.33\textwidth}
\centering
    \includegraphics[width=0.95\linewidth]{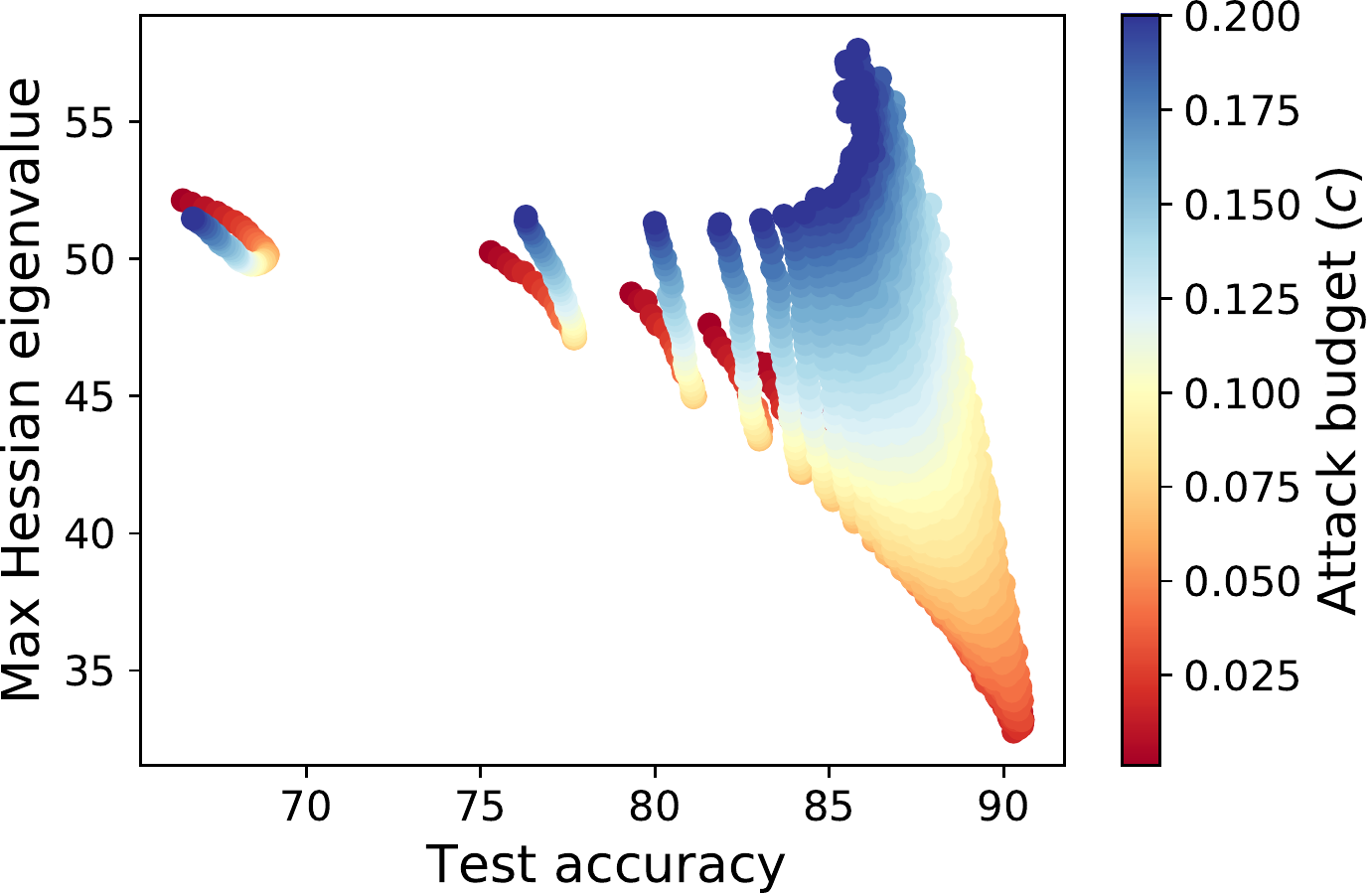}
  \caption{The correspondence between maximum eigenvalue and test accuracy for different training attack budgets ($c$).}
  \label{fig:log_reg_x_acc_eval_y_eigvals_z_attack_epsilon_25056077}
\end{subfigure}%
\begin{subfigure}[t]{.33\textwidth}
\centering
    \includegraphics[width=0.95\linewidth]{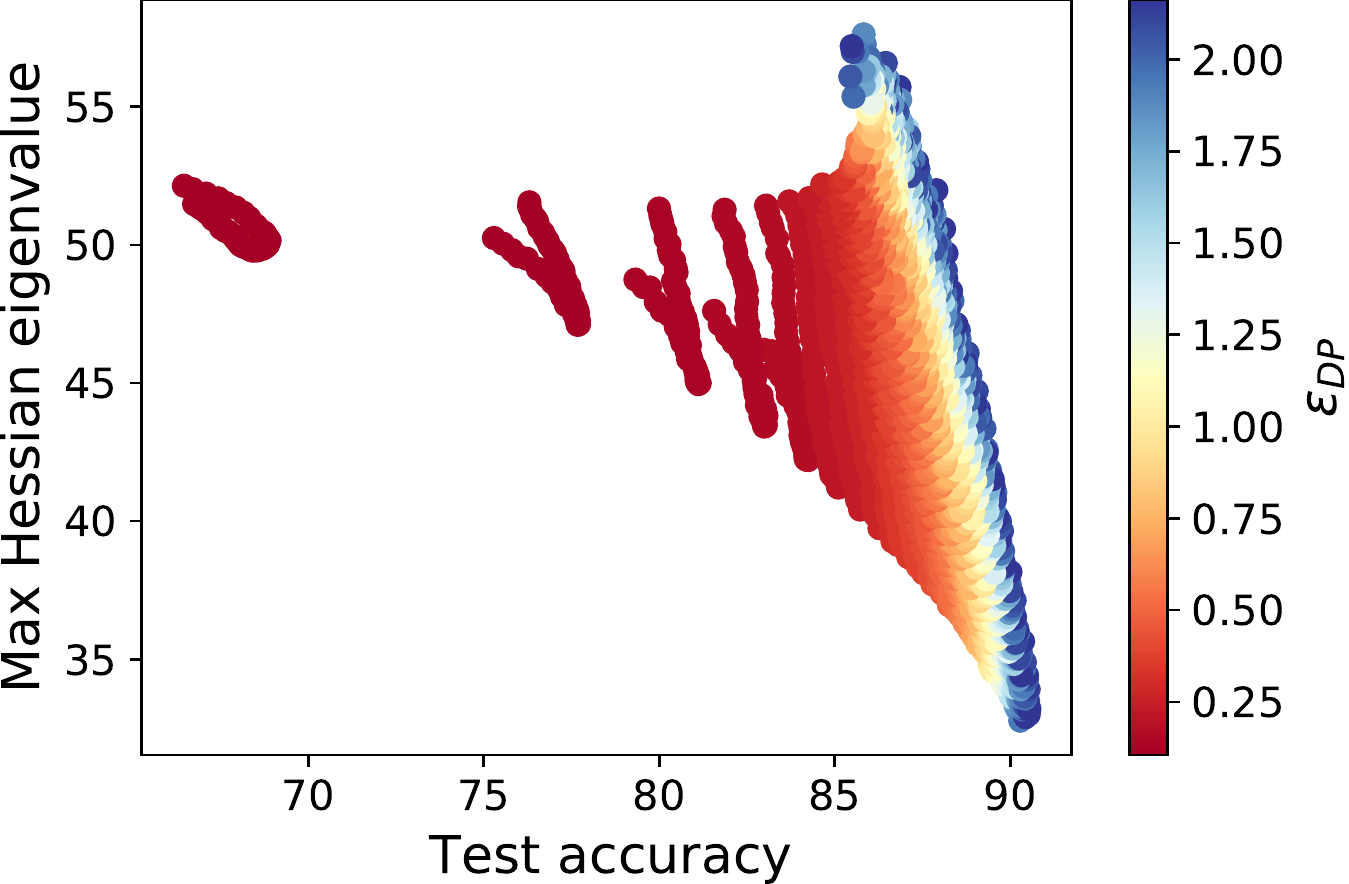}
  \caption{The correspondence between maximum eigenvalue and test accuracy for different privacy guarantees ($\epsilon_{DP}$).}
  \label{fig:log_reg_x_acc_eval_y_eigvals_z_dp_epsilon_25056077}
\end{subfigure}
\caption{Private and robust logistic regression. Each point in the figures represents a different model trained to be robust to perturbations of size $c$ and differentially private with $\epsilon_{DP}$.}
\label{fig:log_reg_25056077}
\end{figure*}

In the previous section we have seen examples of the increased risk brought about by robust and private optimization, and that theory and practice are in alignment. 
We now take a deeper look at why private and robust optimization is difficult.
In particular, we concretize the connection between the smoothness of the loss landscape, adversarial training and the magnitude of learned parameters. Once again, we use the example of binary logistic regression on separable data. 

\paragraph{Adversarial training}

We again consider binary logistic regression with loss  $\ell_a(\theta) = \max_{\norm{\delta}\leq c} \logloss{-y(x+\delta)^T\theta} = \logloss{\advlinlossexample}$. 
We show in \cref{sec: loss_landscape_proofs} that:

\begin{align}
    &\nabla^2 \ell_a(\theta^*) = \frac{c}{2\norm{\theta^*}}\bigg(I - \frac{\theta^*\theta^{*^T}}{\norm{\theta^*}^2}\bigg),
\end{align}
where $\theta^*$ is the optimal solution, perfectly separating data from the two classes.
Clearly $\nabla^2 \ell_a(\theta^*)$ is a positive semi-definite matrix with eigenvalues at $0$ and $\frac{c}{2\norm{\theta^*}}$ -- we can take the eigenvector $v$ to be $\theta^*$ and then $(I - \frac{\theta^*\theta^{*^T}}{\norm{\theta^*}^2})\theta^* = 0$, and we can also take the eigenvector $v$ to be orthogonal to $\theta^*$ resulting in:

\begin{align}
     \frac{c}{2\norm{\theta^*}}\bigg(I - \frac{\theta^*\theta^{*^T}}{\norm{\theta^*}^2}\bigg)v = \frac{c}{2\norm{\theta^*}}v
\end{align}

Thus, the curvature of our loss landscape is completely determined by the ratio of adversarial budget to the magnitude of the optimal parameters, and as $c$ decreases we achieve smoother solutions.

\paragraph{Role of clipping}
We define the clipping operation on a function, $f$, as $clip(f, k) = \clip{f}{k}$, for some $k\in\mathbb{R}_+$.
\citet{xie2018differentially} showed that that gradient clipping implies bounded weight parameters. 
We can assume the optimal parameter takes the form $\theta^*_{clip}:=h(k)\frac{\theta}{\norm{\theta}}$, where $h(k)$ is a monotonically increasing function of the gradient clipping parameter, $k$.
Note, we also assume we can find the optimum under the clipping norm $k>0$, and so we can repeat the above analysis. 
This leaves the smoothness of the loss landscape determined by $\frac{c}{2h(k)}$.
Similar to the purely adversarial setting, this has the intuitive property that decreasing the clipping norm decreases the smoothness of the loss landscape.

\paragraph{Role of noise}
The landscape sharpness is governed by the ratio of adversarial noise to weight norm. 
If we perform noisy gradient descent: $\theta^{t+1} = \theta^{t} - \eta (\nabla + b)$, where $b\sim\mathcal{N}(0, \sigma^2 I_d)$, then $\theta^{t+1}\sim\mathcal{N}(\theta^{t} - \eta\nabla, \eta^2\sigma^2 I_d)$. 
We can use Jensen's inequality to upper bound the expected norm of $\theta^{t+1}$:
\begin{align}
    \E[\norm{\theta^{t+1}}] \leq \E[\norm{\theta^{t+1}}^2]^{\frac{1}{2}} = \sqrt{\norm{\theta^{t} - \eta\nabla}^2 + d\sigma^2}
\end{align}
This implies the expected norm is larger with an additional Gaussian noise term added to gradients.

\paragraph{Empirical validation}
We now demonstrate that these effects are observed in practice using multi-class logistic regression on MNIST. 
We show that smaller adversarial budgets, $c$, and larger clipping norms, $k$, lead to smoother solutions with smaller generalization error. 

Firstly, we train without differentially private noise; only clipping and adversarial training are activated. 
This means our models are not differentially private but allows us to show our analysis holds empirically (c.f. \cref{fig:log_reg_25011747}).
In total, we train 2500 multi-class logistic regression on MNIST, sweeping over different training attack budget ($c$) and clipping threshold ($k$) configurations. 
Each point in the \cref{fig:log_reg_25011747} represents a model trained with a specific ($c$, $k$) configuration.

After this, we will show that moving to differentially private models has virtually no difference empirically, implying our analysis is useful in these settings (c.f. \cref{fig:log_reg_25056077}).
Again, in total, we train 2500 multi-class logistic regression on MNIST, sweeping over different training attack budget ($c$) and differentially private $\epsilon_{DP}$ configurations. 
Each point in the \cref{fig:log_reg_25056077} represents a model trained with a specific ($c$, $\epsilon_{DP}$) configuration.

In \cref{fig:log_reg_x_grad_max_norm_y_attack_epsilon_z_eigvals_25011747}, we plot the maximum eigenvalue of the Hessian for different training attack budgets, $c$, and maximum gradient norm, $k$. 
As expected, the smallest eigenvalues belong to solutions with small $c$ and large $k$.
For values of $k$ close to zero, effective learning becomes increasingly difficult, resulting in non-smooth solutions regardless of the choice of attack budget, $c$. 
Likewise, for attack budgets close to $0.1$, the maximum Hessian eigenvalue remains high even if we increase the clipping threshold, $k$, (i.e. the space of learnable solutions).
In \cref{fig:log_reg_x_dp_epsilon_y_attack_epsilon_z_eigvals_25056077}, we plot analogous results but for fully differential private models, and observe similar results; small attack budgets, $c$, and less privacy (large $\epsilon_{DP}$) results in smaller eigenvalues. 

In \cref{fig:log_reg_x_acc_eval_y_eigvals_z_attack_epsilon_25011747} and \cref{fig:log_reg_x_acc_eval_y_eigvals_z_attack_epsilon_25056077}, we fix $k$ and $\epsilon_{DP}$, respectively, and visualize the relationship between maximum Hessian eigenvalue and test accuracy as a function of the attack budget $c$. 
We can immediately note that smaller eigenvalues (that imply smoother solutions) result in smaller generalization error, confirming that smooth solutions are preferable.
Unsurprisingly, as $c$ increases, the probability of finding a smooth model with high test accuracy decreases.

Similarly, in \cref{fig:log_reg_x_acc_eval_y_eigvals_z_grad_max_norm_25011747} and \cref{fig:log_reg_x_acc_eval_y_eigvals_z_dp_epsilon_25056077}, we fix the attack budget $c$, and visualize the relationship between maximum Hessian eigenvalue and test accuracy as a function of either $k$ or $\epsilon_{DP}$. 
Again, we observe that less smooth solutions are more likely to have smaller test set accuracy.
Decreasing $k$ or $\epsilon_{DP}$ (more privacy) increase the chances of finding a model with a large maximum eigenvalue and (comparatively) small test set accuracy.

We have a theoretical understanding of how clipping and adversarial training affect the learned model. 
Importantly, the trends observed in \cref{fig:log_reg_25011747} are mirrored in \cref{fig:log_reg_25056077}, implying that our analysis is useful in the fully differentially private and robust setting.

\section{Discussion}

Recent work by \citet{song2019privacy} has empirically shown a tension between privacy and robust learning. In particular, it is shown that six state-of-the-art defense methods designed to reduce the success of adversarial examples \emph{increase} the risk of membership inference attacks due to overfitting on the training set \citep{shokri2017membership}.
The phenomenon of overfitting in adversarially robust optimization has also been observed by \citet{rice2020overfitting}. 
If \citet{song2019privacy} showed that robust models are less private, \citet{tursynbek2020robustness} recently identified the contrapositive relation, empirically showing that differentially private models come at the expense of robustness. 
Meanwhile, \citet{ghazi2021robust} shows that the sample complexity of learning both robust and private halfspaces is worse than learning with either property by itself.

\bibliography{main}
\bibliographystyle{icml2021}

\onecolumn
\appendix
\section{Convergence rates of logistic loss on a linearly separable problem with gradient descent}
\label{sec: conv_rates_proofs}

Let $\ell(\cdot, (\cdot, \cdot)):\mathcal{W}\times (\mathcal{X}, \mathcal{Y}) \rightarrow \mathbb{R}$ be the logistic loss, we assume $\mathcal{X}\subseteq \mathbb{R}^d$ and $\forall (x,y)\in(\mathcal{X}, \mathcal{Y})$, $\norm{x}\leq 1$ and $y\in \{\pm1\}$.

Following a similar analysis in \cite{li2019inductive}, we compute convergence rates for gradient descent with and without adversarial training and differential privacy. 
We assume the data is linearly separable with margin $\gamma$, and
set $u=\argmax_{\norm{\theta}=1}\min_{i\in[n]}y_ix_i^T\theta$, the optimal hyperplane that classifies all $(x,y)$ correctly with margin at least $\gamma$.

\noindent \textbf{No privacy / No robustness.} We find convergence rates under the logistic loss $\ell(\theta;y,x)=\logloss{\linloss}$\footnote{We omit the $x$ and $y$ terms in $\ell$ for brevity hereon in.}. The logistic loss has the following first and second derivatives:

\begin{align}
   & \nabla \ell(\theta) = \frac{-ye^{\linloss}x}{1+e^{\linloss}} \\
   & \nabla^2 \ell(\theta) = \frac{e^{\linloss}xx^T}{(1+e^{\linloss})^2} \\
\end{align}

Assuming $\norm{x}\leq 1$, we have $\norm{\nabla \ell(\theta)}\leq 1$ and $\norm{\nabla^2 \ell(\theta)}\leq \frac{1}{4}$. 
We let $L(\theta) = \frac{1}{n}\sum_{i=1}^n \ell(\theta)_i$, and so $\nabla L(\theta) = \frac{1}{n}\sum_{i=1}^n \frac{-y_ie^{\ilinloss}x_i}{1+e^{\ilinloss}}$. 
By Taylor expansion we have:

\begin{align}
    L(\theta^{t+1}) 
    =  L(\theta^t - \eta_t \nabla L(\theta^t)) 
    \leq L(\theta^t) - \eta_t \norm{\nabla L(\theta^t)}^2 + \frac{\eta_t^2}{8} \norm{\nabla L(\theta^t)}^2
\end{align}

where $\eta_t$ is the learning rate at step $t$.
For simplicity, we assume gradient descent is performed with a constant learning rate $\eta_t := \eta$ for every step $t$. 
Our following analysis will depend on a sufficiently small learning rate, which we specify at the relevant places.
Because $L$ is smooth (bounded Hessian), we can apply standard gradient descent convergence analysis. For any $\theta\in\mathbb{R}^d$:

\begin{align}
    \norm{\theta^{t+1} - \theta}^2 &= 
    \norm{\theta^{t} - \theta}^2 - 2\eta\inner{\nabla L(\theta^t)}{\theta^t - \theta} + \eta^2\norm{\nabla L(\theta^t)}^2 \\
    &\leq \norm{\theta^{t} - \theta}^2 - 2\eta(L(\theta^t) - L(\theta)) + \eta^2\norm{\nabla L(\theta^t)}^2 \qquad \text{(by convexity  of $\ell$)} \\
    &\leq \norm{\theta^{t} - \theta}^2 - 2\eta(L(\theta^t) - L(\theta)) + \frac{\eta}{1-\frac{\eta}{8}} (L(\theta^t) - L(\theta^{t+1})) \qquad \text{(by Taylor expansion)} \\
    &= \norm{\theta^{t} - \theta}^2 + (\frac{\eta}{1-\frac{\eta}{8}} -2\eta)L(\theta^{t}) + 2\eta L(\theta) - \frac{\eta}{1-\frac{\eta}{8}} L(\theta^{t+1})
\end{align}

If $\eta < 4$, the contribution of the $L(\theta^{t})$ term is negative and so:

\begin{align}
    \norm{\theta^{t+1} - \theta}^2 
    &\leq \norm{\theta^{t} - \theta}^2 + 2\eta L(\theta) - \frac{\eta}{1-\frac{\eta}{8}} L(\theta^{t+1})
\end{align}

If we set $\theta^0=0$, then $\theta^1 = -\frac{\eta}{n}\sum_{i=1}^n \frac{-y_ix_i}{2}$ and $\norm{\theta^1}\leq\frac{\eta}{2}\leq 1$ as long as $\eta < 2$.
Under the assumption that $L(\theta^{t+1})\leq L(\theta^{t})$, summing the above between $s=1,...,t$, we get:

\begin{align}
    & \norm{\theta^{t+1} - \theta}^2 \leq
    \norm{\theta^{1} - \theta}^2 - t\bigg(\frac{\eta}{1-\frac{\eta}{8}}L(\theta^{t+1}) - 2\eta L(\theta)\bigg) \\
    \implies& L(\theta^{t+1}) \leq \frac{1-\frac{\eta}{8}}{t\eta}(\norm{\theta^1}^2 + \norm{\theta}^2) + 2(1-\frac{\eta}{8})L(\theta)
\end{align}

Since we are free to choose $\theta$, we set it to $\theta:=\frac{\log t}{\gamma}u$, then:

\begin{align}
    L(\theta) &= \avgsum \logloss{\frac{-y\log t}{\gamma}x^Tu}
    \leq \avgsum \logloss{-\log t} 
    = \log(\frac{t+1}{t})
\end{align}

Together with the fact that $\norm{\theta} = \frac{\log t}{\gamma}$, we get the final upper bound:

\begin{align}
    L(\theta^{t+1}) &\leq \frac{1-\frac{\eta}{8}}{t\eta}\bigg(1 + (\frac{\log t}{\gamma})^2\bigg) + 2(1-\frac{\eta}{8})\log(\frac{t+1}{t}) \\
    &= \frac{8-\eta}{8 t\eta}\bigg(1 + (\frac{\log t}{\gamma})^2\bigg) + (\frac{8-\eta}{4})\log(\frac{t+1}{t})
\end{align}

 \noindent \textbf{No privacy / robustness.} We next find convergence rates under gradient-based adversarial training.
 Let $\ell_a(\theta) = \max_{\norm{\delta}\leq c} \logloss{-y(x+\delta)^T\theta} = \logloss{\advlinloss}$. We first find the first and second derivative of this loss.
 
 If we let $\hth := e^{\advlinloss}$, then $\ell_a(\theta) = \log(1+\hth)$ and $\nabla \ell_a(\theta) = \frac{\dhth}{1+\hth}$, where $\dhth =\hth \rth$ if we take $\rth := \rderiv$.
 Similarly:
 
 \begin{align}
    \nabla^2 \ell_a(\theta) = \frac{\hth\rth\rth^T + \hth^2\drth + \hth\drth }{(1+\hth)^2}
\end{align}

To bound the above we must bound $\drth = c(\frac{I}{\norm{\theta}} - \frac{\theta\theta^T}{\norm{\theta}^3})$. 
We first note that the problem can be reduced to finding a lower bound to $\norm{\theta^t}$ for any $t>0$ because $\norm{\drth} \leq \frac{2c}{\norm{\theta}}$.
If we let $\theta^0=0$, then $\ell_a$ is no longer differentiable at zero, so we take the sub-gradient to be:

\begin{align}
    \frac{e^{\advlinloss}(-yx + cr)}{1+e^{\advlinloss}} \in \partial\ell_a(\theta^1)
\end{align}

where $r$ is a vector with $\norm{r}\leq 1$. 
For example, we can take $r=0$, and then the sub-derivative becomes $-\frac{yx}{2}\in \partial\ell_a(\theta^1)$, and is bounded above like $\theta^1 = \gdavgsum{\eta_0}{2}y_ix_i$, and $\inner{\theta^1}{u} \geq \frac{\eta_0 \gamma}{2}$. 
We then take $\eta_t:=\eta$ for $t>1$ such that $\frac{\eta_0}{2} > \eta$.
We then note the inner product between $-\nabla \ell_a(\theta)$ and $u$ is strictly positive:

\begin{align}
    \inner{-\nabla \ell_a(\theta)}{u} \geq \frac{e^{\advlinloss}}{1+e^{\advlinloss}}(\gamma - c) > 0
\end{align}

and so

\begin{align}
    \inner{\theta^2}{u} =  \inner{\theta^1 - \eta \nabla L_a(\theta^1)}{u} 
    =  \inner{\theta^1}{u} + \eta \inner{-\nabla L_a(\theta^1)}{u} 
    \geq \inner{\theta^1}{u} 
    \geq  \eta \gamma  
\end{align}

By a similar argument for any $t>1$, $\inner{\theta^t}{u} \geq \eta \gamma$. 
Because $u$ is the global minimizer it follows that $\norm{\theta^t} \geq \eta \gamma$.
 We have a lower bound on $\theta^t$ for any $t$ and so $\norm{\drth} \leq \frac{2c}{\eta \gamma}$.
 
 The other term we upper bound is $\rth\rth^T=xx^T -2cyx^T\frac{\theta}{\norm{\theta}} + c^2\frac{\theta\theta^T}{\norm{\theta}^2}$, and since we assume bounded data we can bound this like $\norm{\rth\rth^T} \leq (1+c)^2$.
 Piecing this altogether gives:
 
 \begin{align}
    \norm{\nabla^2L_a(\theta)} &\leq \norm{\frac{\hth\rth\rth^T}{(1+\hth)^2}}  + \norm{\frac{\hth(1+\hth)\drth}{(1+\hth)^2}}\\
    &= \norm{\frac{\hth\rth\rth^T}{(1+\hth)^2}}  + \norm{\frac{\hth\drth}{(1+\hth)}}\\
    &\leq \norm{\frac{\hth}{(1+\hth)^2}}\norm{\rth\rth^T}  + \norm{\frac{\hth}{(1+\hth)}}\norm{\drth} \\
    &\leq \frac{(1+c)^2}{4} + \frac{2c}{\eta\gamma}
\end{align}
 
 Letting $s:=\frac{(1+c)^2}{4} + \frac{2c}{\eta\gamma}$,
 now that we have a bound on the second derivatives we can again use the Taylor expansion of gradient descent:
 
 \begin{align}
    L_a(\theta^{t+1}) &=  L_a(\theta^t - \eta \nabla L_a(\theta^t)) \\
    &\leq L_a(\theta^t) - \eta \norm{\nabla L_a(\theta^t)}^2 + \frac{s\eta^2}{2} \norm{\nabla L_a(\theta^t)}^2
\end{align}

Then for any $\theta\in\mathbb{R}^d$, 

\begin{align}
    \norm{\theta^{t+1} - \theta}^2 
    &\leq \norm{\theta^{t} - \theta}^2 - 2\eta(L_a(\theta^t) - L_a(\theta)) + \frac{\eta}{1-\frac{s\eta}{2}} (L_a(\theta^t) - L_a(\theta^{t+1})) \\
\end{align}

As long as $\eta$ satisfies $s\eta<1$, the $L_a(\theta^t)$ contribution is negative and we get:

\begin{align}
    \norm{\theta^{t+1} - \theta}^2 
    &\leq \norm{\theta^{t} - \theta}^2 + 2\eta L_a(\theta) - \frac{2\eta}{2-s\eta} L_a(\theta^{t+1}) \\
\end{align}

Isolating $\eta$ in the condition $s\eta<1$ gives $\eta<\frac{4(\gamma-2c)}{\gamma(1+c)^2}$ which also implies that we require $c<\frac{\gamma}{2}$ for our results to hold.
Summing over the iterations as before gives:

\begin{align}
    & \norm{\theta^{t+1} - \theta}^2 \leq
    \norm{\theta^{1} - \theta}^2 - t(\frac{2\eta}{2-s\eta}L_a(\theta^{t+1}) - 2\eta L_a(\theta)) \\
    \implies& \frac{2t\eta}{2-s\eta}L_a(\theta^{t+1}) \leq \norm{\theta^{1} - \theta}^2 + 2t\eta L_a(\theta)
\end{align}

By choosing $\theta = \frac{\log t}{\gamma - c}u$ we again get $\ell_a(\theta)=\log(1+\frac{1}{t})$ and $\norm{\theta} = \frac{\log t}{\gamma - c}$, giving the final upper bound:

\begin{align}
L_a(\theta^{t+1}) &\leq \frac{2-s\eta}{2t\eta}(\norm{\theta}^2 + \norm{\theta^1}^2 + 2t\eta L_a(\theta)) \\
&\leq \frac{2-s\eta}{2t\eta}\bigg((\frac{\log t}{\gamma - c})^2 + (1+c)^2\bigg) + (2-s\eta)\log(1+\frac{1}{t})
\end{align}
 
\noindent \textbf{Privacy / No robustness.} We next find convergence rates under gradient descent with differential privacy. 
At each step we add noise; we do not clip gradients as they are already $1$-Lipschitz and we assume here the clipping norm is larger than one. 
Let $\overline{\nabla L(\theta)} = \nabla L(\theta) + b$, where $b\sim\mathcal{N}(0, \sigma^2I_d)$. 
Throughout the following analysis we find convergence results under expectations taken over the noise added for differential privacy.

Our analysis is almost identical to the non-private case, and we sometimes write $\nabla$ or $\overline{\nabla}$ where the context is clear.
Firstly, the Taylor expansion follows:

\begin{align}
    L(\theta^{t+1}) &=  L(\theta^t - \eta \overline{\nabla}) \\
   &\leq L(\theta^t) - \eta \nabla^T\overline{\nabla}  + \frac{\eta^2}{2}\overline{\nabla}^T\nabla^2\overline{\nabla}
\end{align}

Note that $\norm{\nabla^2}\leq \frac{1}{4}$ and $\overline{\nabla} = \nabla + b$. Although we take expectations over the random noise in following, we omit the notation $\E$ for conciseness:

\begin{align}
    L(\theta^{t+1})
   &\leq L(\theta^t) - \eta \norm{\nabla}^2 + \frac{\eta^2}{8}(\norm{\nabla}^2 + d\sigma^2)
\end{align}

Then for any $\theta\in\mathbb{R}^d$, and again taking expectations over the noise,

\begin{align}
    \norm{\theta^{t+1} - \theta}^2 
    &= \norm{\theta^{t} - \theta}^2 - 2\eta\inner{\nabla + b}{\theta^t - \theta} + \eta^2\norm{\nabla + b}^2 \\
    &\leq \norm{\theta^{t} - \theta}^2 + 2\eta L(\theta) - \frac{8\eta}{8-\eta} L(\theta^{t+1}) + \frac{8d\eta^2\sigma^2}{8-\eta}
\end{align}

 Repeating the same process as in the non-private case we get:
 
 \begin{align}
    L(\theta^{t+1}) &\leq 
    \frac{8-\eta}{8 t\eta}\bigg(1 + d\sigma^2 + (\frac{\log t}{\gamma})^2\bigg) + (\frac{8-\eta}{4})\log(\frac{t+1}{t}) +  \eta d\sigma^2
\end{align}
 
 \noindent \textbf{Privacy / robustness.} We can repeat the analysis as in non-private and robust case, while taking expectations over the random noise to give the upper bound:
 
 \begin{align}
L_a(\theta^{t+1}) 
&\leq \frac{2-s\eta}{2t\eta}\bigg( (1+c)^2 + d\sigma^2 + (\frac{\log t}{\gamma - c})^2 \bigg) + (2-s\eta)\log(1+\frac{1}{t}) +  \eta d\sigma^2
\end{align}
 
Importantly, if we compare the non-private case to the private case for either the robust or non-robust setting, we take an equivalent hit in convergence that depends on the variance of noise and dimensionality of data.

\paragraph{Empirical adversarial risk comparison}
\Cref{fig:lr_example} is not an apples-to-apples comparison because we are comparing standard loss ($L$) under standard training against robust loss ($L_a$) under adversarial training. 
We can compare with robust accuracy under standard training by noting that under gradient descent, if $\norm{\theta^1}\leq 1$, then $\norm{\theta^t}\leq 1 + \eta(t-1)$, and together with: 

\begin{align}
 \ell_a(\theta) &= \logloss{\advlinlossexample} \\
 &= \log(e^{-c\norm{\theta}_*} + e^{\linloss}) + c\norm{\theta}_* \\ 
 &\leq \log(1 + e^{\linloss}) + c\norm{\theta}_* \\
 &= \ell(\theta) + c\norm{\theta}_*,
\end{align}

we get the following upper bound on the robust loss under gradient descent:

\begin{align}
\begin{split}
 L_a(\theta^t) \leq  &\frac{8-\eta}{8 t\eta}\bigg(1 + (\frac{\log t}{\gamma})^2\bigg) + \\ &(\frac{8-\eta}{4})\log(\frac{t+1}{t}) + c\big(1 + \eta(t-1)\big)
\end{split}
\end{align}

\section{Binary logistic regression with clipping and adversarial training.}
\label{sec: loss_landscape_proofs}

The following analysis only considers a single input, but can extended to a batch of inputs through simple averaging.
We again consider binary logistic regression with loss  $\ell_a(\theta) = \max_{\norm{\delta}\leq c} \logloss{-y(x+\delta)^T\theta} = \logloss{\advlinloss}$. 
As stated previously the first and second derivatives are given by:

\begin{align}
    &\nabla \ell_a(\theta) = \frac{\hth \rth}{1+\hth} \\
    &\nabla^2 \ell_a(\theta) = \frac{\hth\rth\rth^T + \hth^2\drth + \hth\drth }{(1+\hth)^2}
\end{align}
 
where $\hth := e^{\advlinloss}$, $\rth := \rderiv$, and $\drth = c(\frac{I}{\norm{\theta}} - \frac{\theta\theta^T}{\norm{\theta}^3})$.
At the optimal $\theta^*=\argmin_{\theta} \ell_a(\theta)$, we have $\rderivopt=0 \implies \theta^* = \frac{yx\norm{\theta^*}}{c}$. The $\hth\rth\rth^T$ term in the second derivative vanishes at the optimum and $\hthopt=1$ and we are left with:

\begin{align}
    &\nabla^2 \ell_a(\theta^*) = \frac{\drthopt}{2} = \frac{c}{2\norm{\theta^*}}\bigg(I - \frac{\theta^*\theta^{*^T}}{\norm{\theta^*}^2}\bigg)
\end{align}
 
Clearly $\nabla^2 \ell_a(\theta^*)$ is a positive semi-definite matrix with eigenvalues at $0$ and $\frac{c}{2\norm{\theta^*}}$ -- we can take the eigenvector $v$ to be $\theta^*$ and then $(I - \frac{\theta^*\theta^{*^T}}{\norm{\theta^*}^2})\theta^* = 0$, and we can also take the eigenvector $v$ to be orthogonal to $\theta^*$ resulting in:

\begin{align}
     \frac{c}{2\norm{\theta^*}}\bigg(I - \frac{\theta^*\theta^{*^T}}{\norm{\theta^*}^2}\bigg)v = \frac{c}{2\norm{\theta^*}}v
\end{align}

Thus, the curvature of our loss landscape is completely determined by the ratio of adversarial noise to the magnitude of the optimal parameters, and as $c$ decreases we achieve smoother solutions.

\section{Excess risk analysis for robust \& private learning}

So far we have concentrated our analysis on logistic regression.
We now give a simple risk analysis for general convex losses for robust learning with differential privacy.

Let $\ell(\cdot, (\cdot, \cdot)):\mathcal{W}\times (\mathcal{X}, \mathcal{Y}) \rightarrow \mathbb{R}$ be a loss function with the following properties: $L$-Lipschitz, $G$-smooth, convex and linear in its parameters, $\ell(w, (x, y)) = \ell(yw^Tx)$, $\forall w\in\mathcal{W}$, $\norm{w}\leq \beta$, $\mathcal{X}\subseteq \mathbb{R}^d$, $\forall (x,y)\in(\mathcal{X}, \mathcal{Y})$, $\norm{x}\leq 1$ and $y\in \{\pm1\}$.

\noindent Consider two batch updates that differ in their final element:
\begin{align}
    &P = \sum_{i=1}^{n-1} y_i\nabla\ell(y_iw_t^T x_i)x_i + y_n\nabla\ell(y_nw_t^T x_n)x_n + z \\
    &Q = \sum_{i=1}^{n-1} y_i\nabla\ell(y_iw_t^T x_i)x_i + y_n'\nabla\ell(y_n'w_t^T x_n')x_n' + z
\end{align}

\noindent Setting $A=\sum_{i=1}^{n-1} y_i\nabla\ell(y_iw_t^T x_i)x_i$, $B=y_n\nabla\ell(y_nw_t^T x_n)x_n$, $B'=y_n'\nabla\ell(y_n'w_t^T x_n')x_n'$, we have $P=\mathcal{N}(A+B, \sigma^2 I_d)$ and $Q=\mathcal{N}(A+B', \sigma^2 I_d)$. To achieve $(\epsilon, \delta)-DP$, we must bound $\alpha_M(\lambda) = \max_{X, X'}\alpha_M(\lambda; X, X')$ as defined in \cite{abadi2016deep}, where $\alpha_M(\lambda)=\lambda D_{\lambda + 1}(P || Q)$ is as defined in \cite{bun2016concentrated}. This results in the following:

\begin{align}
    \alpha_{M_k}(\lambda) =\lambda D_{\lambda + 1}(P \rVert Q)=\frac{\lambda(\lambda + 1)\norm{B-B'}^2}{2\sigma^2}
\end{align}

\noindent Under the assumption that $\norm{x}\leq 1$, $y\in \{\pm1\}$, and $\ell$ is L-Lipschitz, we have the following upper bound:

\begin{align}
    \norm{B-B'} \leq 2L \implies \alpha_{M_k}(\lambda) \leq \frac{2\lambda(\lambda + 1)L^2}{\sigma^2}
\end{align}

\noindent Summing over all iterations gives:

\begin{align}
    \alpha_M(\lambda) \leq \sum_{k=1}^T \alpha_{M_k}(\lambda) \leq \frac{2\lambda(\lambda + 1)L^2T}{\sigma^2} \leq \frac{c\lambda^2L^2T}{\sigma^2}
\end{align}

\noindent for some constant $c$. Then for some $c'$, taking $\sigma^2 = \frac{c'L^2T\log(\frac{1}{\delta})}{\epsilon^2}$ will give $(\epsilon, \delta)-DP$ according to Theorem 2.2 in \cite{abadi2016deep}.

\noindent Turning to the robust setting, we wish to solve $\min_{w\in\mathcal{W}}\max_{\norm{v}_p \leq r} \ell\big(yw^T(x+v)\big)$, and in the convex setting the inner-maximization can be given in closed-form as:

\begin{align}
    \max_{\norm{v}_p \leq r} \ell\big(yw^T(x+v)\big) = \ell(yw^Tx - r\norm{w}_q)
\end{align}

\noindent where $\frac{1}{p} + \frac{1}{q}=1$. If we set $p=\infty$, then squared Euclidean distance between $B$ and $B'$ becomes

\begin{align}
    \norm{B-B'} &\leq \norm{y_n\nabla\ell_nx_n - y_n'\nabla\ell_n'x_n' - r\sign(w)\nabla\ell_n + r\sign(w)\nabla\ell_n'} \\
    &\leq 2\sqrt{d}(1+r)L
\end{align}

\noindent This similarly holds for $p=2$ by noting that that the derivative of $\norm{w}_2$ is of unit length and replacing $\sign{(w)}$ with this unit length value. Then $(\epsilon, \delta)-DP$ holds by taking $\sigma^2 = \frac{c'(1+r)^2L^2T\log(\frac{1}{\delta})}{\epsilon^2}$.

First note that if $\ell$ is $L$-Lipschitz, then $\mathbb{E}[\norm{\hat{g}}^2]= \norm{g}^2 + \mathbb{E}[\norm{\hat{z}}^2]\leq L^2 + d\sigma^2$.
Assuming $\ell$ is $\lambda$-strongly convex, then by Theorem 1 of \cite{shamir2013stochastic} the bound on excess risk is given by the following:

\begin{align}
    \mathbb{E}[\ell(w_{T}) - \ell(w^{*})] \leq \frac{17(L^2 + d\sigma^2)(1+\log(T))}{\lambda T}
\end{align}

where for $t>0$ the per step learning rate satisfies $\eta_t=\frac{1}{\lambda t}$.

\noindent For private learning we can plug in $\sigma^2 = \frac{c'L^2T\log(\frac{1}{\delta})}{\epsilon^2}$ and for robust and private learning we can plug in $\sigma^2 = \frac{c'(1+r)^2L^2T\log(\frac{1}{\delta})}{\epsilon^2}$.

\end{document}